\documentclass[letterpaper, 10 pt, journal, twoside]{IEEEtran}
\usepackage{graphicx}
\usepackage{amsmath}
\usepackage{amssymb}
\usepackage{cite}
\usepackage{array}
\usepackage{booktabs}
\usepackage{multirow}
\usepackage{makecell}
\usepackage{balance}

\newcolumntype{M}[1]{>{\centering\arraybackslash}m{#1}}

\newcommand{\argmin}{\operatornamewithlimits{argmin}}

\begin{document}
\title{Ellipse Regression with Predicted Uncertainties for \\Accurate Multi-View 3D Object Estimation}

% \author{Michael~Shell,~\IEEEmembership{Member,~IEEE,}
%         John~Doe,~\IEEEmembership{Fellow,~OSA,}
%         and~Jane~Doe,~\IEEEmembership{Life~Fellow,~IEEE}% <-this % stops a space
% \thanks{M. Shell was with the Department
% of Electrical and Computer Engineering, Georgia Institute of Technology, Atlanta,
% GA, 30332 USA e-mail: (see http://www.michaelshell.org/contact.html).}% <-this % stops a space
% \thanks{J. Doe and J. Doe are with Anonymous University.}% <-this % stops a space
% \thanks{Manuscript received April 19, 2005; revised August 26, 2015.}}
\author{Wenbo~Dong,~\IEEEmembership{Student~Member,~IEEE,}
%		Pravakar~Roy,~\IEEEmembership{Student~Member,~IEEE,}
		and~Volkan~Isler$^{1}$,~\IEEEmembership{Senior~Member,~IEEE}%
%\thanks{Manuscript received: Month, Day, Year; Revised Month, Day, Year; Accepted Month, Day, Year.}%Use only for final RAL version
%\thanks{This paper was recommended for publication by Editor FirstName A. EditorName upon evaluation of the Associate Editor and Reviewers' comments.
%This work was supported by (organizations/grants which supported the work.)} %Use only for final RAL version
\thanks{$^{1}$W. Dong, and V. Isler are with the Department of Computer Science and Engineering, University of Minnesota, Minneapolis, MN, 55455, USA (e-mail: dongx358@umn.edu, isler@cs.umn.edu).}%
\thanks{This work was supported in part by USDA NIFA MIN-98-G02, in part by the MnDrive initiative, and in part by NSF \#1722310.}
%\thanks{Manuscript received February 24, 2020.}
%\thanks{Digital Object Identifier (DOI): see top of this page.}
}
\maketitle

\begin{abstract}
Convolutional neural network (CNN) based architectures, such as Mask R-CNN, constitute the state of the art in object detection and segmentation.
Recently, these methods have been extended for model-based segmentation where the network outputs the parameters of a geometric model (e.g. an ellipse) directly.
This work considers objects whose three-dimensional models can be represented as ellipsoids.
We present a variant of Mask R-CNN for estimating the parameters of ellipsoidal objects by segmenting each object and accurately regressing the parameters of projection ellipses.
We show that model regression is sensitive to the underlying occlusion scenario and that prediction quality for each object needs to be characterized individually for accurate 3D object estimation.
We present a novel ellipse regression loss which can learn the offset parameters with their uncertainties and quantify the overall geometric quality of detection for each ellipse.
These values, in turn, allow us to fuse multi-view detections to obtain 3D ellipsoid parameters in a principled fashion.
The experiments on both synthetic and real datasets quantitatively demonstrate the high accuracy of our proposed method in estimating 3D objects under heavy occlusions compared to previous state-of-the-art methods.
%Recent CNN-based object detectors (e.g., Ellipse R-CNN) have been designed to represent and infer occluded objects as ellipses.
%However, such models do not predict uncertainties for regressed parameters in different occlusion scenarios, in which the prediction qualities need to be characterized for accurate 3D object estimation.
%In this letter, we propose a novel ellipse regression loss for learning ellipse offset parameters along with their uncertainties and predicting the geometric quality for each detection in 2D.
%The learned uncertainties and detection quality allow us to integrate multi-view detections into a probabilistic framework, which further improves the performance of localizing occluded objects as ellipsoids in 3D.
%The experiments on both synthetic and real datasets quantitatively demonstrate the highest accuracy of our proposed method in estimating 3D objects from heavy occlusions compared to previous state-of-the-art methods.
\end{abstract}


% \fi
\IEEEpeerreviewmaketitle

%----------------------------------------------------------------------
% SECTION I: Introduction
%----------------------------------------------------------------------
\section{Introduction} \label{sec:introduction}
\begin{figure}[t]
	\centering
	\includegraphics[width=0.99\columnwidth]{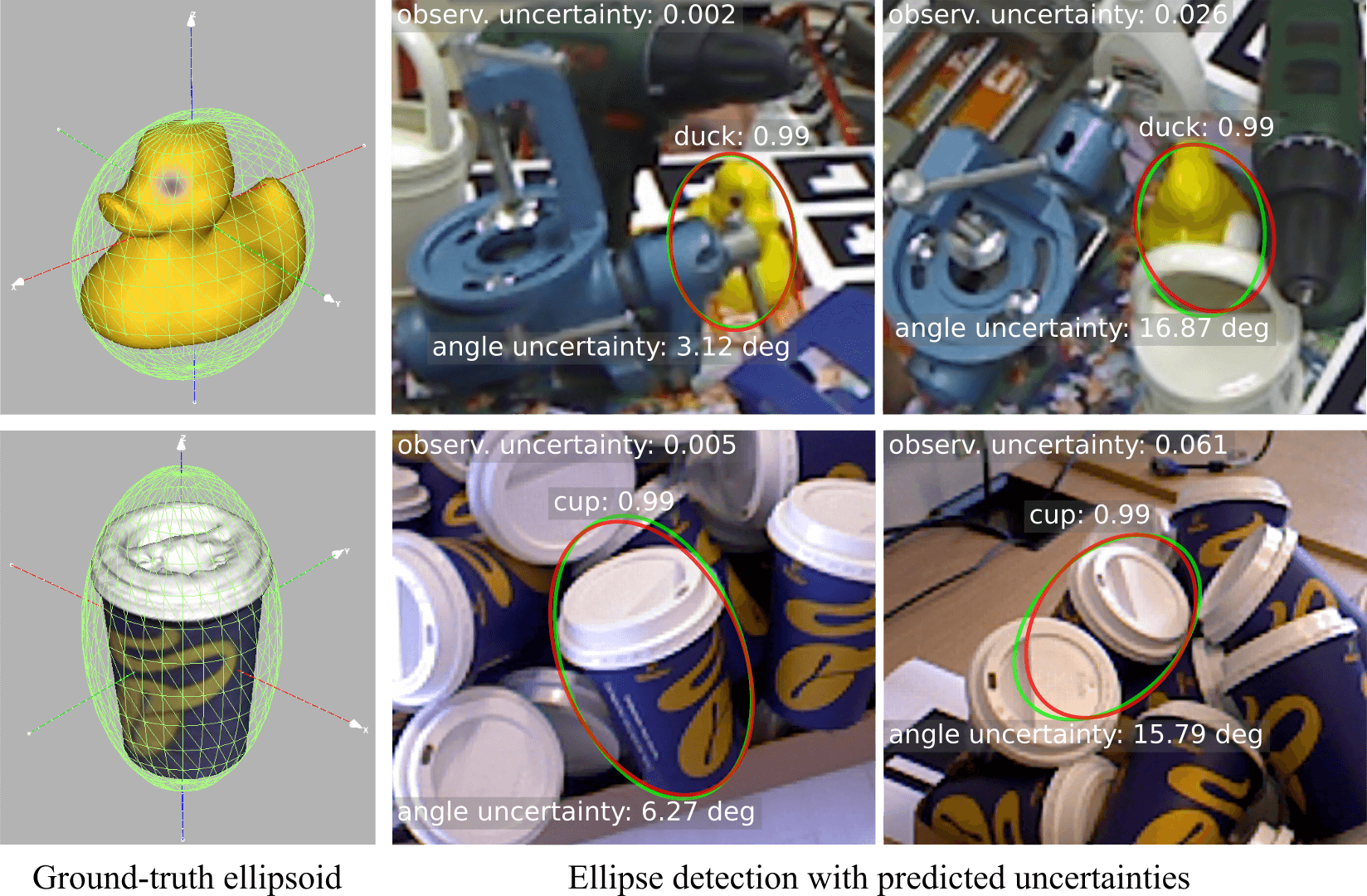}
	\caption{Our proposed model predicts the observation uncertainty for each ellipse detection and the uncertainties for the corresponding ellipse parameters based on different occlusion levels. Left: The ground truth (GT) of 3D enclosing ellipsoids (green) for Duck~\cite{hinterstoisser2012model, brachmann2014learning} and Cup~\cite{doumanoglou2016recovering} datasets. Right: All four ellipse detections (red) have the same classification score, while the rightmost two have higher uncertainties with heavily occluded regions (the GT ellipses are colored green). The geometric quality for each detected object is characterized by its observation uncertainty. The ellipse-angle uncertainty is better visualized from the image (the uncertainties of the other ellipse parameters are demonstrated in Sec.~\ref{sec:experiments}).}
	\label{fig:motivUncertain}
\end{figure}

\begin{figure}[!t]
	\centering
	\includegraphics[width=0.99\columnwidth]{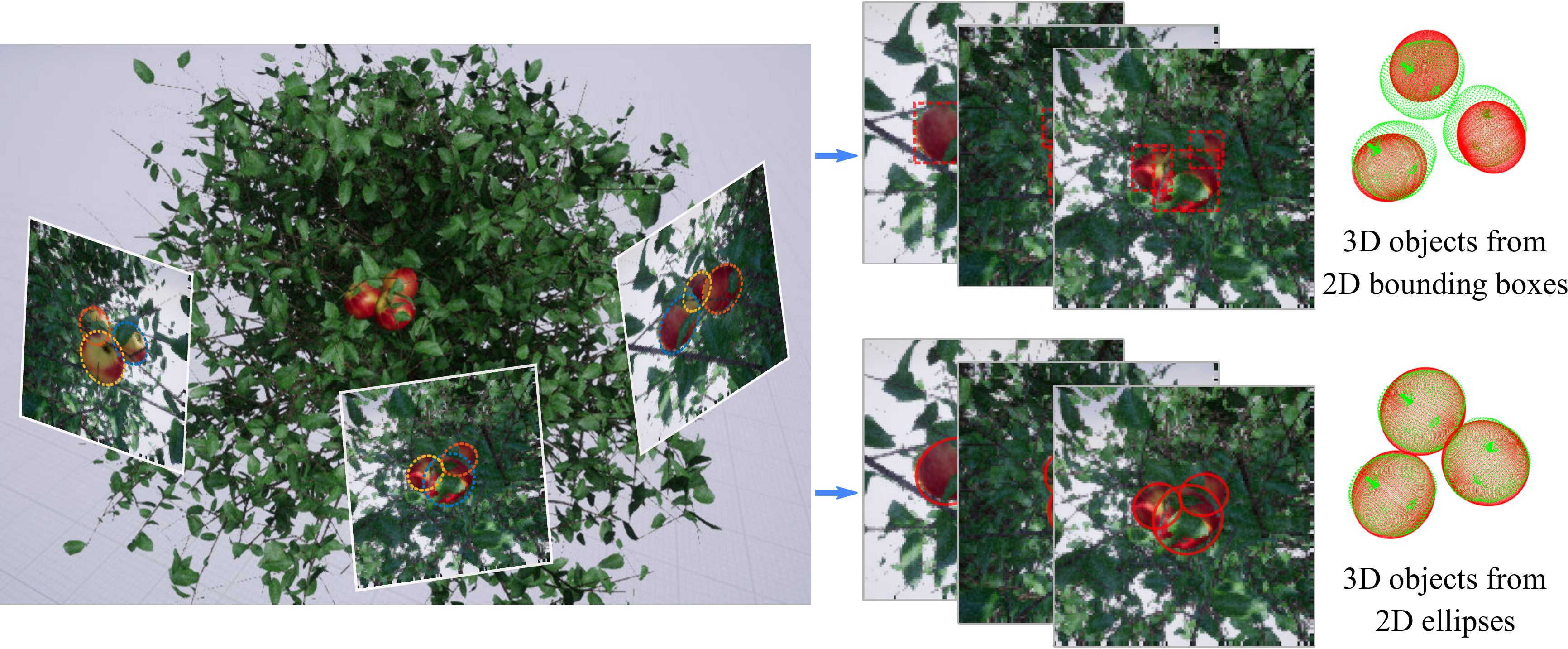}
	\caption{3D estimation of occluded objects from multi-view image detections. Left: A cluster of occluded objects captured from multiple views. Right: Multi-view 3D object estimation~\cite{rubino20183d} from ellipse detections~\cite{dong2020ellipse} outputs much more accurate sizes in 3D and poses in 6D than using bounding-box constraints~\cite{nicholson2019quadricslam} (3D GT points are colored green).}
	\label{fig:motivation3D}
\end{figure}

% for IEEE journal papers produced under \LaTeX\ using
% IEEEtran.cls version 1.8b and later.
\IEEEPARstart{D}{etection} and 3D localization of heavily occluded objects in cluttered scenes, such as fruit clusters in trees~\cite{dong2020ellipse}, is a hard problem as objects often vary significantly in pose and appearance with occlusion (see Fig.~\ref{fig:motivUncertain}).
Adapting convolutional neural networks (CNNs) for object detection~\cite{ren2017faster} and instance segmentation (e.g., Mask R-CNN~\cite{he2018mask}) to this canonical task is a promising way to extract object information in 2D.
To better infer the entire object shape from heavy occlusions, the authors in~\cite{dong2020ellipse} proposes the Ellipse R-CNN model that focuses on the visible regions and detects objects as ellipses.
Moreover, 2D ellipse detections are suitable inputs for 3D localization and size estimation of ellipsoid objects, while the bounding boxes are unreliable in terms of 2D object representation due to insufficient geometric constraints from multiple views~\cite{nicholson2019quadricslam} (see Fig.~\ref{fig:motivation3D}).

Object detection in CNN-based models above is typically formulated as a regression problem, which outputs an object region and its classification score per prediction.
For example, Ellipse R-CNN relies on ellipse regression to localize occluded objects in 2D.
However, we observe that the regression accuracy is relatively low when the object is heavily occluded, which makes it hard to accurately estimate the object size and pose in 3D.
It implies that uncertainties should be considered in the regression to characterize different occlusion levels, while the traditional loss for regression (i.e., the smooth $\text{L}_1$ loss~\cite{girshick2015fast}) does not take such ambiguities into account.
Besides, ellipse detection with a high classification score is assumed to have low uncertainties, which is not always true (see Fig.~\ref{fig:motivUncertain}).

To address the above problems, we propose a novel ellipse regression that predicts the uncertainty for each learned ellipse parameter and the observation uncertainty for the entire detected ellipse.
Specifically, the uncertainties of ellipse parameters indicate the occlusion level for the detected object and how much each regressed parameter is affected by the occlusion.
Furthermore, the geometric quality of the detected ellipse is captured by the observation uncertainty, which determines how good this detection is and how much it weighs in the multi-view 3D object estimation to reduce the localization error.

The primary contributions of this letter are twofold:
\begin{itemize}
	\item We propose to formulate the ellipse-regression loss as KL divergence for learning uncertainties of ellipse parameters and observation uncertainty.
	Each ellipse-parameter prediction and its ground truth (GT) are modeled as Gaussian distribution and Dirac delta function, respectively.
	For the observation uncertainty, we treat the predicted and GT ellipses as two Gaussian distributions.
	The regression loss is thus defined as the KL divergence of the prediction and GT distributions.
	Different from previous work~\cite{he2019bounding} learning absolute values in pixels, we predict uncertainties for ellipse offsets (with visible regions as the reference~\cite{dong2020ellipse}), such that the learned values are not sensitive to different object sizes and image resolutions.
	\item To estimate occluded objects in 3D, we first parameterize the object landmarks as Quadrics~\cite{rubino20183d}.
	Then we develop a probabilistic framework that integrates the uncertainties of each view (weighted differently) into the multi-view object estimation system.
	By taking into account the view equality, our model accurately estimates the 3D enclosing ellipsoids of detected objects from heavy occlusions.
\end{itemize}

To demonstrate the generality of our uncertainty model, we evaluate the extended Ellipse R-CNN model on synthetic and real datasets of various occluded objects, and illustrate how the proposed method helps improve the accuracy of 3D object estimation from occlusion.

%----------------------------------------------------------------------
% SECTION II: Related Work
%----------------------------------------------------------------------
\section{Related Work} \label{sec:relatedWork}
Our goal is to estimate the 3D object size and pose using detection uncertainties from multiple images that exhibit severe occlusions due to other nearby objects.
In the following section, we discuss recent work on CNN-based object detectors, uncertainty modeling, and 3D
object estimation.

\subsection{CNN-Based Object Detectors}
Recently, great improvements in object detection tasks on Pascal~\cite{everingham2010pascal}, ImageNet~\cite{krizhevsky2012imagenet}, and MS COCO datasets~\cite{lin2014microsoft} have been achieved and attributed to the successful development of CNNs that are categorized into single-shot~\cite{redmon2016you, liu2016ssd} and R-CNN~\cite{ren2017faster, he2018mask, girshick2015fast} architectures.
Although single-shot detection algorithms are efficient, R-CNN approaches by integrating region proposal and classification into two stages, have greatly improved the accuracy, and are currently the state-of-the-art object detectors.
However, it is still challenging to accurately detect occluded objects, since the detection performance drops significantly as objects cluster and occlude each other~\cite{dollar2011pedestrian, dong2020ellipse}.
One of the most common strategies of occlusion handling is either learning pre-defined semantic parts~\cite{tian2015deep} or guiding attention on visible features~\cite{hu2018squeeze} given a bounding-box region.
Recent work~\cite{dong2020ellipse} proposes a detector that is trained on visible regions to infer the entire object as an ellipse, which highly reduces false positives due to the feature similarity within object clusters.
This gives us a promising way to further localize occluded objects in 3D from their 2D ellipse detections.

\subsection{Uncertainty Modeling in Neural Networks}
Bayesian neural networks (BNNs)~\cite{neal2012bayesian} have demonstrated that a Bayesian model can be integrated into neural networks to obtain the uncertainty of prediction.
In practice, BNNs use dropout connections in forward passes during the inference stage to generate variations in prediction~\cite{gal2016dropout}.
The uncertainty is thus modeled by such variations in the form of distribution.
This approximation strategy based on Monte Carlo samples is further developed in Bayesian SegNet~\cite{kendall2015bayesian} to yield pixel-wise uncertainties in semantic segmentation results.
However, it is hard to determine the optimal configuration of dropout layers for different learning architectures.
Besides, Monte Carlo sampling requires additional inference time.
The KL loss in~\cite{he2019bounding} is formulated to predict uncertainties in bounding-box regression, but the predicted values are in the absolute image scale and not stable for the objects with largely different sizes.
In contrast, our method exploits the KL loss to learn ellipse uncertainties that are normalized based on visible regions to treat equally the objects with arbitrary sizes, orientations and occlusions.
The KL loss enables us to learn ellipse regression and uncertainty prediction at the same time.

\subsection{3D Object Estimation from 2D Detections}
Modeling objects as quadrics from 2D detections has been investigated in recent research~\cite{crocco2016structure}, and been further developed in~\cite{rubino20183d, nicholson2019quadricslam} for estimating 3D object pose and size from CNN-based detectors.
However, the entire shape of a heavily occluded object is hardly retrieved from a bounding box that captures a little portion of the visible part (see Fig.~\ref{fig:motivation3D}).
While some efforts have been made for 3D localization of elliptical objects~\cite{roy2018registering, dong2020semantic}, by applying the standard mapping techniques~\cite{wu2013towards, mur2017orb}, none of them is able to estimate the object size because of low-resolution 3D reconstructions.
Moreover, all these works represent visible object regions as bounding boxes, which are not appropriate to serve as inputs for further estimating the object pose and size due to the spatial ambiguities in the rectangle constraints of an ellipsoid.
Specifically, it is not robust to fit an inscribed ellipse for each bounding box~\cite{rubino20183d}, since there may exist infinite solutions of ellipses that all satisfy the bounding-box constraints~\cite{nicholson2019quadricslam} even from multiple views.
We thus develop the Ellipse R-CNN model to detect occluded objects as ellipses and predict their uncertainties for accurate 3D object estimation.

%----------------------------------------------------------------------
% SECTION III: Proposed Uncertainty Prediction
%----------------------------------------------------------------------
\section{Proposed Ellipse Uncertainty Prediction} \label{sec:uncertainty}
In this section, we first introduce the ellipse regression in Ellipse R-CNN.
We then propose to use the KL divergence as the regression loss for learning ellipse uncertainties that are normalized with respect to the visible object region.

\begin{figure*}[!t]
	\centering
	\includegraphics[width=0.99\textwidth]{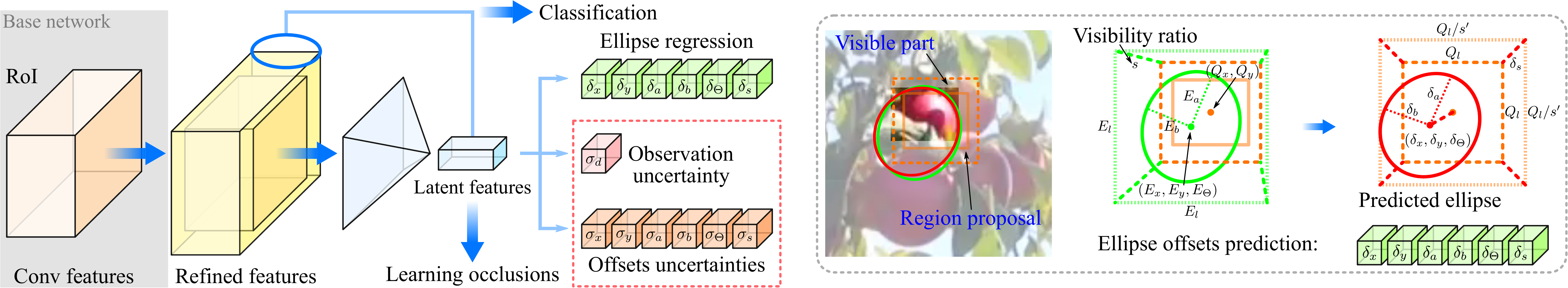}
	\caption{Left: Our network architecture for estimating ellipse uncertainties (observation uncertainty and offsets uncertainties) along with the ellipse regression in occluded cases. The uncertainties are modeled as standard deviations that are taken into account in the KL loss. Right: Overview of ellipse regression in the Ellipse R-CNN model~\cite{dong2020ellipse}. The regressor predicts the offset $\delta_s$ to estimate the visibility ratio $s$ between the extended square $Q$ (dashed brown square of length $Q_l$) of the proposed visible part (solid brown box) and the enclosing square of the GT ellipse $E$ (dashed green box).}
	\label{fig:learnUncertainty}
\end{figure*}

\subsection{Overview of Ellipse Regression} \label{subsec:overveiwEllipse}
To effectively reduce false positives in heavily occluded scenarios, the Ellipse R-CNN model~\cite{dong2020ellipse} focuses on the visible object region to regress five ellipse parameters (see Fig.~\ref{fig:learnUncertainty}).
Given a proposal of the visible region $P = (P_x, P_y, P_w, P_h)$ with its GT ellipse $E = (E_x, E_y, E_a, E_b, E_{\Theta})$, we extend $P$ as the square $Q = (Q_x, Q_y, Q_l)$ sharing the same center $(P_x, P_y)$ with its length as $Q_l = \sqrt{{P_w}^2 + {P_h}^2}$ to avoid distorting the ellipse orientation $E_{\Theta} \in (-\pi/2, \pi/2]$.
We then parameterize the regression in terms of six offsets $\delta_x$, $\delta_y$, $\delta_a$, $\delta_b$, $\delta_{\Theta}$ and $\delta_s$ to predict the ellipse as $E^{\prime}$:
\begin{equation} \label{eq:ellipseRegScale}
\begin{aligned}
& \delta_x = s^{\prime}(E^{\prime}_x - Q_x) / Q_l , \quad \delta_y = s^{\prime}(E^{\prime}_y - Q_y) / Q_l , \\
& \delta_a = \log (2s^{\prime}E^{\prime}_a / Q_l) , \quad \delta_b = \log (2s^{\prime}E^{\prime}_b / Q_l) , \\
& \delta_s = \log \left((s^{\prime} + 1) / 2\right) , \quad \delta_{\Theta} = E^{\prime}_{\Theta} / \pi , \\
& \delta^{\ast}_x = s(E_x - Q_x) / Q_l , \quad \delta^{\ast}_y = s(E_y - Q_y) / Q_l , \\
& \delta^{\ast}_a = \log (2sE_a / Q_l) , \quad \delta^{\ast}_b = \log (2sE_b / Q_l) , \\
& \delta^{\ast}_s = \log \left((s + 1) / 2\right) , \quad \delta^{\ast}_{\Theta} = E_{\Theta} / \pi ,
\end{aligned}
\end{equation}
where $s = Q_l / E_l$, $s \in (0, 1]$ characterizes the visibility calculated as the ratio between the size $Q_l$ of the extended square $Q$ (of the visible part) and the length $E_l = 2\sqrt{{E_a}^2 + {E_b}^2}$ of the square enclosing the ellipse $E$ (i.e., the entire object region).
By predicting $s^{\prime}$, we transfer the offset reference from the visible part $Q_l$ to the entire object region $Q_l / s^{\prime}$.
It guarantees that, as the proposal $P$ is located close to the small visible region, all predicted values $\delta$ (with the target $\delta^{\ast}$) are normalized with bounded magnitudes even in heavily occluded cases (e.g., $s \rightarrow 0$ when $Q_l \rightarrow 0$).
For a fully visible object, $s = 1$ and $\delta^{\ast}_s = 0$, which indicates that Eq.~\eqref{eq:ellipseRegScale} is a generalized ellipse regression that can deal with both occluded and unoccluded cases.

Predicting relative offsets instead of absolute ellipse parameters has two key benefits:
(1) Since the extended square $Q$ is proportional to the ellipse size, all six predicted offsets are normalized such that the objects with hugely different sizes and arbitrary orientations weight equally in the total regression loss to make the learning process unaffected by absolute pixel values.
(2) The normalization guarantees that the prediction values are all close to zero (with small magnitudes) when the proposed region $P$ is near the GT visible part, which stabilizes the training process without outputting unbounded values.

\subsection{Learning Uncertainties with KL Loss}
Our key idea for learning ellipse uncertainties is to estimate the probability distributions of the whole predicted ellipse and its six offsets that are normalized based on the visible region.
Specifically, we assume that the predicted ellipse offsets $\delta$ are independent variables and each parameter can be modeled as the univariate Gaussian distribution:
\begin{equation} \label{eq:likeSingle}
\mathcal{P}_{\mathbf{w}}(z) = \dfrac{1}{\sqrt{2\pi \sigma^2}} \exp \Big( -\dfrac{(z - z_p)^2}{2\sigma^2} \Big) ,
\end{equation}
where $\mathbf{w}$ denotes the learnable weights of the network, and $z_p$ is one of the predicted ellipse offsets.
The standard deviations $\sigma$ (i.e., $\sigma_x$, $\sigma_y$, $\sigma_a$, $\sigma_b$, $\sigma_{\Theta}$ and $\sigma_s$) measure the uncertainties of offsets estimation.
As $\sigma \rightarrow 0$, it infers that the network is more confident about the predicted offsets.
Each of the GT ellipse offsets can be treated as a special Gaussian distribution with $\sigma$ being 0, which is a Dirac delta function:
\begin{equation} \label{eq:diracSingle}
\mathcal{P}_{\text{D}}(z) = \Delta (z - z_g) ,
\end{equation}
where $z_g$ is one of the GT offset parameters.

\begin{figure}[!t]
	\centering
	\includegraphics[width=0.99\columnwidth]{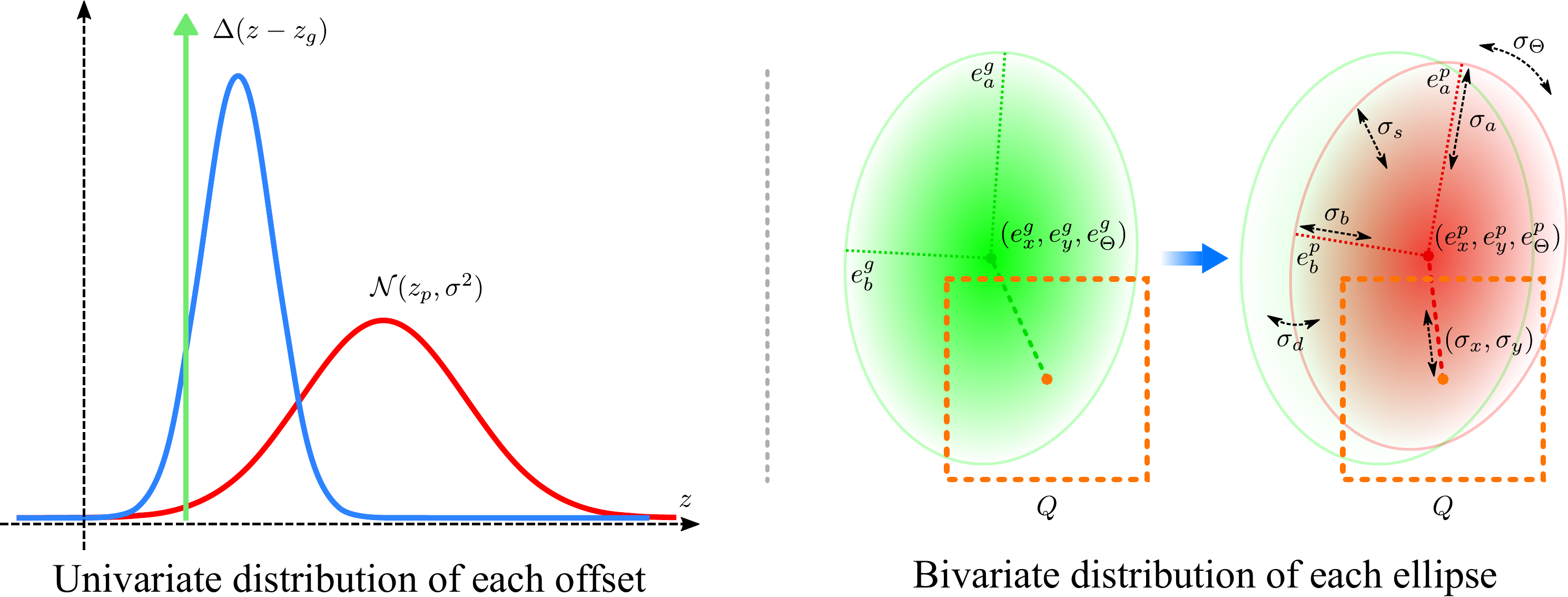}
	\caption{Left: Two univariate Gaussian distributions (blue and red) are the predictions of an ellipse offset, and the Dirac delta function (green) is the distribution of the GT offset. The red one is less accurate and thus has a larger uncertainty $\sigma$. Right: The predicted (red) and GT (green) ellipses normalized from the extended square $Q$ are two bivariate Gaussian distributions. Seven uncertainties characterize the prediction quality of $e^{p}$.}
	\label{fig:probDistribution}
\end{figure}

To predict the ellipse offsets and uncertainties at the same time, we minimize the KL divergence (namely the KL loss) between $\mathcal{P}_{\mathbf{w}}(z)$ and $\mathcal{P}_{\text{D}}(z)$~\cite{robert2014machine} for each training sample:
%\begin{equation} \label{eq:minKLdiv}
%\mathcal{L}_{\text{off}} = \dfrac{1}{N}\sum D_{\text{KL}} \left( \mathcal{P}_{\text{D}}(z) \| \mathcal{P}_{\mathbf{w}}(z) \right) .
%\end{equation}
%The KL loss for a single sample can be further simplified as:
\begin{equation} \label{eq:offLoss}
\begin{aligned}
\mathcal{L}_{\text{off}} &= D_{\text{KL}} \left( \mathcal{P}_{\text{D}}(z) \| \mathcal{P}_{\mathbf{w}}(z) \right) \\
&= \int \mathcal{P}_{\text{D}}(z) \log \mathcal{P}_{\text{D}}(z) dz - \int \mathcal{P}_{\text{D}}(z) \log \mathcal{P}_{\mathbf{w}}(z) dz \\
&= \dfrac{(z_g - z_p)^2}{2\sigma^2} + \dfrac{1}{2} \log (\sigma^2) + \dfrac{1}{2} \log(2\pi) + \mathcal{F}(\mathcal{P}_{\text{D}}(z)) \\
&\propto \dfrac{(z_g - z_p)^2}{2\sigma^2} + \dfrac{1}{2} \log (\sigma^2) ,
\end{aligned}
\end{equation}
where $\tfrac{1}{2} \log(2\pi)$ and $\mathcal{F}(\mathcal{P}_{\text{D}}(z))$ are ignored since they do not depend on the estimated $\mathbf{w}$.
For the orientation offset $\delta_{\Theta}$, $z_g - z_p$ is rectified as $\rho (\delta^{\ast}_{\Theta}, \delta_{\Theta})$ so that the angle difference $\varphi = \pi (\delta^{\ast}_{\Theta} - \delta_{\Theta})$ is within $(-\pi/2, \pi/2]$ to handle critical angles~\cite{dong2020ellipse} (e.g., the error between $\pi/2$ and $-\pi/2$):
\begin{equation} \label{eq:ellipseRegLoss}
\begin{aligned}
\rho (\delta^{\ast}_{\Theta}, \delta_{\Theta}) =
\begin{cases}
{\rm atan2}\left(\sin\varphi, \cos\varphi\right) & \text{if } \cos\varphi \ge 0 \\
{\rm atan2}\left(-\sin\varphi, -\cos\varphi\right) & \text{if } \cos\varphi < 0
\end{cases} .
\end{aligned}
\end{equation}
The KL loss $D_{\text{KL}}$ in Eq.~\eqref{eq:offLoss} is a generalized loss: the squared loss $\tfrac{1}{2} (z_g - z_p)^2$ is the special case as $\sigma = 1$.
This enables the network to further minimize the loss by predicting larger variance $\sigma^2$ when the parameter $z_p$ is estimated inaccurately away from $z_g$ (see. Fig.~\ref{fig:probDistribution}).
To avoid loss exploding in Eq.~\eqref{eq:offLoss} due to the small values of $\sigma$ at the early stage of training, we predict $\alpha = \log (\sigma^2)$ instead of $\sigma$:
\begin{equation} \label{eq:propMod}
\mathcal{L}_{\text{off}} =
\begin{cases}
\tfrac{1}{2} \exp(-\alpha) \left(z_g - z_p\right)^2 + \tfrac{1}{2}\alpha & \text{if } |z_g - z_p| \le 1 \\
\exp(-\alpha) \left(|z_g - z_p| - \tfrac{1}{2}\right) + \tfrac{1}{2}\alpha & \text{if } |z_g - z_p| > 1
\end{cases} ,
\end{equation}
where the smooth $\mathcal{L}_1$ loss~\cite{girshick2015fast} is applied to the term $z_g - z_p$.

\textbf{Observation Uncertainty:}
To evaluate the geometric quality of the entire predicted ellipse, we propose to predict one more uncertainty $\sigma_d$ that can be used to weight each prediction accordingly in the 3D object estimation.
Specifically, an ellipse in 2D $(x, y, a, b, \Theta)$ is treated as the bivariate Gaussian distribution $\mathcal{N}(\boldsymbol\mu, \boldsymbol\Sigma)$ with $\boldsymbol\mu = \big(\begin{smallmatrix} x\\y \end{smallmatrix}\big)$ and $\boldsymbol\Sigma = \mathbf{T} \boldsymbol \Lambda \mathbf{T}^{-1}$, where $\mathbf{T} = \big(\begin{smallmatrix} \cos\Theta & -\sin\Theta\\ \sin\Theta & \cos\Theta \end{smallmatrix}\big)$ and $\boldsymbol \Lambda = \operatorname{diag} (a^2, b^2)$.
To maintain the benefits described in Sec.~\ref{subsec:overveiwEllipse}, we exploit normalized ellipses based on the offsets in Eq.~\eqref{eq:ellipseRegScale}:
\begin{equation} \label{eq:normEllipse}
\begin{aligned}
& e^{p}_x = \delta_x / s^{\prime}, \ \ e^{p}_y = \delta_y / s^{\prime} , \ \ s^{\prime} = 2\exp (\delta_s) - 1 , \\
& e^{p}_a = \tfrac{1}{2} \exp (\delta_a) / s^{\prime} , \ \ e^{p}_b = \tfrac{1}{2} \exp (\delta_b) / s^{\prime} , \ \ e^{p}_{\Theta} = \pi \delta_{\Theta} , \\
& e^{g}_x = \delta^{\ast}_x / s , \ \ e^{g}_y = \delta^{\ast}_y / s , \\
& e^{g}_a = \tfrac{1}{2} \exp (\delta^{\ast}_a) / s , \ \ e^{g}_b = \tfrac{1}{2} \exp (\delta^{\ast}_b) / s , \ \ e^{g}_{\Theta} = \pi \delta^{\ast}_{\Theta} ,
\end{aligned}
\end{equation}
where $e^{p}$ and $e^{g}$ are the predicted and GT ellipses normalized from the extended square $Q$, respectively.

We model the KL divergence $D_{\text{KL}}$ between an ellipse $e$ and $e^{p}$ as the univariate Gaussian distribution, and model the $D_{\text{KL}}$ between $e$ and $e^{g}$ as the Dirac delta function:
\begin{equation} \label{eq:likeEllipse}
\mathcal{P}_{\mathbf{w}}(d) = \dfrac{1}{\sqrt{2\pi \sigma_d^2}} \exp \Big( -\dfrac{(e - e^{p})^2}{2\sigma_d^2} \Big) , \ \ \mathcal{P}_{\text{D}}(d) = \Delta (e - e^{g}) ,
\end{equation}
where $D_{\text{KL}} \left( e \| e^{p} \right)$ and $D_{\text{KL}} \left( e \| e^{g} \right)$ are described as $e - e^{p}$ and $e - e^{g}$ from a distance perspective.
Using the same strategies in Eq.~\eqref{eq:offLoss} and~\eqref{eq:propMod}, we obtain the KL loss $\mathcal{L}_{\text{obs}}$ to estimate $\alpha_d$:
\begin{equation} \label{eq:obsLoss}
\begin{aligned}
\mathcal{L}_{\text{obs}} &= D_{\text{KL}} \left( \mathcal{P}_{\text{D}}(d) \| \mathcal{P}_{\mathbf{w}}(d) \right) \\
&\propto \exp(-\alpha_d) \cdot \mathcal{L}_1(e_g - e_p) + \tfrac{1}{2} \alpha_d ,
\end{aligned}
\end{equation}
where $\alpha_d = \log (\sigma_d^2)$ and $e_g - e_p$ denotes the KL divergence between $e_g$ and $e_p$: $D_{\text{KL}} \left( \mathcal{N}_g \| \mathcal{N}_p \right) = \tfrac{1}{2} \operatorname{tr}(\boldsymbol\Sigma_p^{-1}\boldsymbol\Sigma_g) + \tfrac{1}{2} (\boldsymbol\mu_p - \boldsymbol\mu_g)^{\mathsf{T}} \boldsymbol\Sigma_p^{-1} (\boldsymbol\mu_p - \boldsymbol\mu_g) -1 + \tfrac{1}{2} \ln \big( \tfrac{\det\boldsymbol\Sigma_p}{\det\boldsymbol\Sigma_g} \big)$, as illustrated in Fig.~\ref{fig:probDistribution}.

The six ellipse offsets and seven uncertainties are generated by multilayer perceptrons (MLPs)~\cite{bishop1995neural} on top of the latent features encoded from the refined features (see Fig.~\ref{fig:learnUncertainty}).
To learn ellipse regression with predicted uncertainties, we estimate $\hat{\mathbf{w}}$ that minimizes the total regression loss $\mathcal{L}_{\text{reg}}$ over $N$ samples:
\begin{equation} \label{eq:minKLdiv}
\begin{aligned}
\hat{\mathbf{w}} &= \argmin_{\mathbf{w}} \dfrac{1}{N} \sum_i \Big(\sum_{j \in \star} \mathcal{L}_{\text{off}} + \mathcal{L}_{\text{obs}} \Big) , \\
\star &= \{x, y, a, b, \Theta, s\} .
\end{aligned}
\end{equation}

%----------------------------------------------------------------------
% SECTION IV: Probabilistic 3D Object Estimation
%----------------------------------------------------------------------
\section{Probabilistic 3D Object Estimation}

\subsection{Sensor Models based on Dual Quadrics}
Quadrics are surfaces in 3D (e.g., ellipsoids) that are represented by a $4\times4$ symmetric matrix $\mathbf{Q}$, and conics $\mathbf{C}$ are the 2D counterparts of $\mathbf{Q}$ (e.g., ellipses).
The dual form $\mathbf{Q}^{\ast}$ (the adjoint of $\mathbf{Q}$) can be projected onto an image plane to create the dual conic $\mathbf{C}^{\ast} = \mathbf{P} \mathbf{Q}^{\ast} \mathbf{P}^{\mathsf{T}}$,
where $\mathbf{P} = \mathbf{K} [\mathbf{R} | \mathbf{t}]$ is the projection matrix that contains
intrinsic ($\mathbf{K}$) and extrinsic camera parameters.
The dual quadrics can be parameterized as $\mathbf{Q}^{\ast} = \mathbf{Z} \breve{\mathbf{Q}}^{\ast} \mathbf{Z}^{\mathsf{T}}$ (see~\cite{rubino20183d}). Here, $\breve{\mathbf{Q}}^{\ast} = \operatorname{diag} (s_1^2, s_2^2, s_3^2, -1)$ denotes an ellipsoid centered at the origin, and $\mathbf{Z} = \big(\begin{smallmatrix} \mathbf{R}(\boldsymbol\theta) & \mathbf{t} \\ \textbf{0}_{1\times3} & 1 \end{smallmatrix}\big)$ is a homogeneous transformation, where $\mathbf{t} = (t_1, t_2, t_3)$ is the translation of the quadric centroid, the angles $\boldsymbol\theta = (\theta_1, \theta_2, \theta_3)$ define the rotation matrix, and the shape of the ellipsoid along its three semi-axes is $\mathbf{s} = (s_1, s_2, s_3)$.
We compactly represent $\mathbf{Q}^{\ast}$ as a vector $\mathbf{q} = (\theta_1, \theta_2, \theta_3, t_1, t_2, t_3, s_1, s_2, s_3)^{\mathsf{T}}$.
%we refer the reader to textbooks on projective geometry~\cite{hartley2003multiple};

Each detected object is represented by the predicted ellipse offsets $\mathbf{e} = (\delta_x, \delta_y, \delta_a, \delta_b, \delta_{\Theta}, \delta_s)$ and the square region $\boldsymbol Q = (Q_x, Q_y, Q_l)$ as the reference.
%The normalized ellipse is represented as $\mathbf{e} = (e^{p}_x, e^{p}_y, e^{p}_a, e^{p}_b, e^{p}_{\Theta})$;
Given the camera pose $\mathbf{T}_i$, the estimated object $\mathbf{q}_j$ and the detected offsets $\mathbf{e}_{ij}$, the estimated offsets $\hat{\mathbf{e}}_{ij}$ and the estimated divergence $\hat{d}_{ij}$ are defined as:
\begin{equation} \label{eq:sensors}
\begin{aligned}
\boldsymbol\tau(\mathbf{T}_i, \mathbf{q}_j) &= \operatorname{off} \Big(\operatorname{ellipse} (\mathbf{P} \mathbf{Q}_{(\mathbf{q}_j)}^{\ast} \mathbf{P}^{\mathsf{T}}), \boldsymbol Q_{ij} \Big) = \hat{\mathbf{e}}_{ij} , \\
\boldsymbol\kappa(\mathbf{e}_{ij}, \mathbf{T}_i, \mathbf{q}_j) &= \operatorname{div} \Big(\operatorname{ellipse} (\mathbf{P} \mathbf{Q}_{(\mathbf{q}_j)}^{\ast} \mathbf{P}^{\mathsf{T}}), \boldsymbol Q_{ij}, \mathbf{e}_{ij} \Big) = \hat{d}_{ij} ,
 \end{aligned}
\end{equation}
where the operator $\operatorname{ellipse} (\cdot)$ takes the adjoint of the projected conic and solves for the five ellipse parameters~\cite{spain2007analytical}.
$\operatorname{off} (\cdot)$ and $\operatorname{div} (\cdot)$ calculate the ellipse offsets and ellipse divergence based on Eq.~\eqref{eq:ellipseRegScale} and~\eqref{eq:likeEllipse}, respectively.
We assume that the data association is solved~\cite{agarwal2013robust} and given.
%The reference square $\boldsymbol Q_{ij}$ (being known given $i$ and $j$) is omitted, since

\subsection{Probabilistic Model for Multi-view 3D Object Estimation}
The conditional probability over all camera poses $\mathcal{T} = \{\mathbf{T}_i\}$ and objects $\mathcal{Q} = \{\mathbf{q}_j\}$ given the ellipse divergence $\mathcal{D} = \{d_{ij}\}$ and ellipse offsets $\mathcal{E} = \{\mathbf{e}_{ij}\}$ is modeled and factored based on the Bayes Theorem:
\begin{equation} \label{eq:prob}
\mathcal{P}(\mathcal{Q}, \mathcal{T} | \mathcal{D}, \mathcal{E}) = \dfrac{\mathcal{P}(\mathcal{D} | \mathcal{E}, \mathcal{Q}, \mathcal{T}) \cdot \mathcal{P}(\mathcal{E} | \mathcal{Q}, \mathcal{T}) \cdot \mathcal{P}(\mathcal{Q}, \mathcal{T})}{\mathcal{P}(\mathcal{D}, \mathcal{E})} .
\end{equation}
To estimate the objects in 3D from multiple views, we perform the maximum a posteriori (MAP) estimation.
Since the denominator $\mathcal{P}(\mathcal{D}, \mathcal{E})$ is constant, we ignore this normalization factor. In addition, $\mathcal{P}(\mathcal{Q}, \mathcal{T})$ is ignored as it is a uniform distribution without any prior information.
Essentially, we can optimize $\mathcal{Q}$ and $\mathcal{T}$ by minimizing the negative log the joint probability:
\begin{equation} \label{eq:minProb}
\mathcal{Q}^{\ast}, \mathcal{T}^{\ast} = \argmin_{\mathcal{Q}, \mathcal{T}} - \log \big( \mathcal{P}(\mathcal{D} | \mathcal{E}, \mathcal{Q}, \mathcal{T}) \cdot \mathcal{P}(\mathcal{E} | \mathcal{Q}, \mathcal{T}) \big) .
\end{equation}

At the inference stage, the prediction $\alpha$ is converted back to $\sigma$.
The likelihood $\mathcal{P}(\mathcal{E} | \mathcal{Q}, \mathcal{T})$ is thus modeled as a Gaussian $\mathcal{N}(\boldsymbol\tau(\mathbf{T}_i, \mathbf{q}_j), \boldsymbol\Lambda_{ij})$, where $\boldsymbol\Lambda = \operatorname{diag}(\sigma_x^2, \sigma_y^2, \sigma_a^2, \sigma_b^2, \sigma_{\Theta}^2, \sigma_s^2)$ is the covariance matrix that captures the uncertainties of each predicted offset.
The factor $\mathcal{P}(\mathcal{D} | \mathcal{E}, \mathcal{Q}, \mathcal{T})$ is also assumed to be a Gaussian $\mathcal{N}(\boldsymbol\kappa(\mathbf{e}_{ij}, \mathbf{T}_i, \mathbf{q}_j), \lambda_{ij})$ with the variance $\lambda = \sigma_d^2$ that characterizes the uncertainty for the geometric quality of the entire ellipse.
We can further formulate the problem in Eq.~\eqref{eq:minProb} as a nonlinear least-squares optimization:
\begin{equation} \label{eq:minLS}
\argmin_{\mathcal{Q}} \sum_{ij} \Big( \| \boldsymbol\kappa(\mathbf{e}_{ij}, \mathbf{T}_i, \mathbf{q}_j) \|^2_{\lambda_{ij}} + \| \mathbf{e}_{ij} - \boldsymbol\tau(\mathbf{T}_i, \mathbf{q}_j) \|^2_{\boldsymbol\Lambda_{ij}} \Big) .
\end{equation}
In this letter, the camera poses $\mathcal{T}$ are assumed to be known for 3D object estimation.
We obtain the initial object parameters for optimization by using the closed-form solution from multi-view ellipse detections~\cite{rubino20183d}.

%----------------------------------------------------------------------
% SECTION V: Experiments
%----------------------------------------------------------------------
\section{Experiments} \label{sec:experiments}
To demonstrate our proposed uncertainty model for accurate 3D object estimation, we conduct experiments using synthetic and real-world object images, and evaluate the quality of the estimated objects both quantitatively and qualitatively.

\begin{figure}[!t]
	\centering
	\includegraphics[width=0.99\columnwidth]{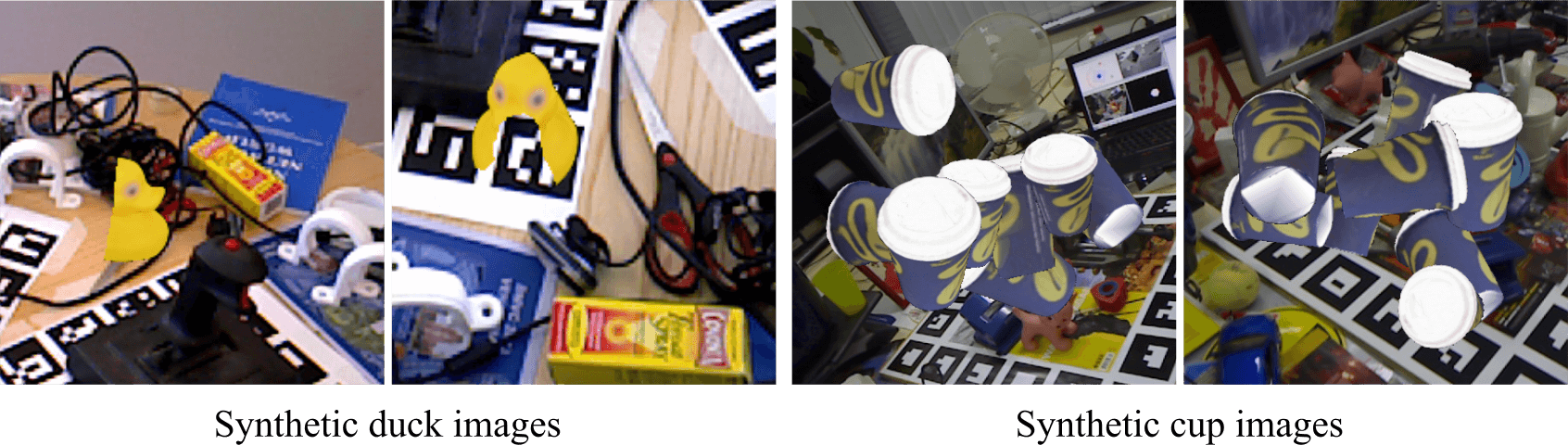}
	\caption{Synthetic images of the Duck and Cup datasets as the training data. To add interference, the image backgrounds are filled with real images from two datasets\cite{tejani2014latent, hinterstoisser2012model}, respectively. Each object on the image is further augmented with random scaling, shifting, and rotation during training~\cite{sundermeyer2018implicit}.}
	\label{fig:trainingData}
\end{figure}

\subsection{Datasets}
We first evaluate the detection accuracy of our Ellipse R-CNN+ (i.e., predicting ellipse parameters and uncertainties) on the synthetic occluded fruits (SOF) dataset~\cite{dong2020ellipse}.
By comparing to the state-of-the-art techniques, we further demonstrate how our probabilistic model based on the uncertainties improves the accuracy of estimating 3D occluded objects on two public datasets: Duck~\cite{hinterstoisser2012model, brachmann2014learning} and Cup~\cite{doumanoglou2016recovering} datasets.
The SOF dataset consists of 3,545 images (3,040 for training and 505 for testing) of fruit clusters occluded within a realistic tree (the visibility ratio $r_o \ge 0.3$). We generate the images by varying the 3D poses and sizes of each fruit model in Unreal Engine (UE), and replace the background with random images taken from different real orchards~\cite{dong2020semantic}.
The GT ellipses and visible object regions are obtained by projecting the 3D fruit ellipsoids onto the corresponding images~\cite{qiu2017unrealcv} based on known camera poses.

The Duck dataset is built upon the Hinterstoisser~\cite{hinterstoisser2012model} dataset, where we select the image sequence of a duck toy occluded by other objects with the GT object poses provided by the Brachmann~\cite{brachmann2014learning} dataset.
For the training data, we render 1,313 views of the duck model based on the viewing poses in the Hinterstoisser dataset.
The background is randomly filled by the Tejani dataset~\cite{tejani2014latent}.
From each view, we also render three to five different objects (e.g., vise, driller, and cat) at their random poses to partially block the duck (with the visibility ratio $r_o \in [0.3, 0.6]$), and we only keep its visible part (see Fig.~\ref{fig:trainingData}).
For testing, we select 102 images of heavily occluded duck ($r_o \in [0.3, 0.6]$) in a real office environment to evaluate the proposed uncertainty prediction and probabilistic 3D object estimation.
The Cup dataset is built in a similar way.
In the Doumanoglou dataset~\cite{doumanoglou2016recovering}, we render 2,377 views of a cluster of occluded cups ($r_o \ge 0.3$) at random poses for training.
The image background is randomly replaced by the Hinterstoisser dataset (see Fig.~\ref{fig:trainingData}).
The testing data includes two sequences of cup clusters that have 56 images of 19 cups and 61 images of 16 cups on a table, respectively.
For both duck and cup models, we calculate their minimum volume enclosing ellipsoids~\cite{moshtagh2005minimum}  whose image projections are obtained as the GT ellipses based on the GT camera poses (see Fig.~\ref{fig:motivUncertain}).

\begin{figure}[!t]
	\centering
	\includegraphics[width=0.99\columnwidth]{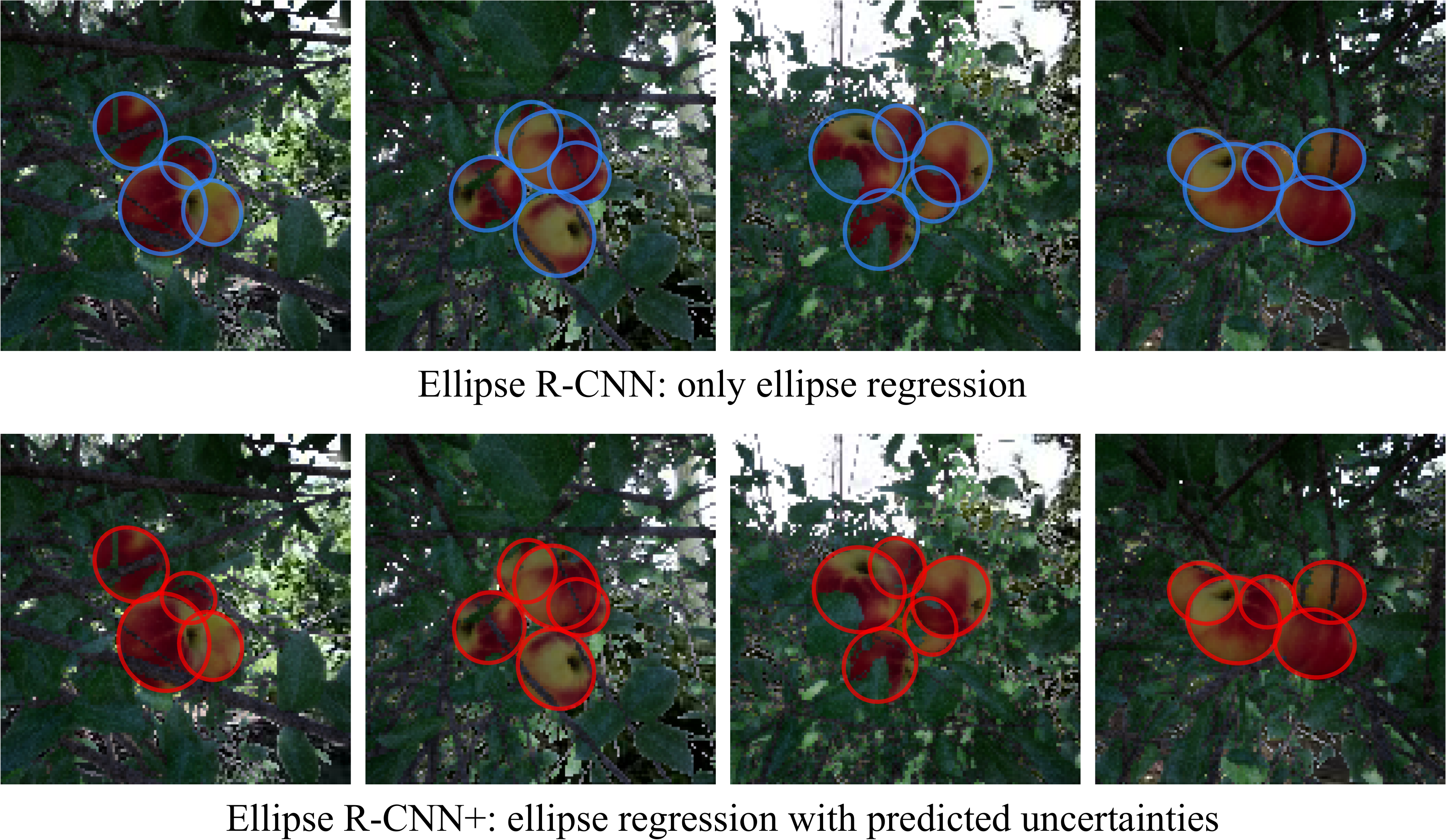}
	\caption{Qualitative results of object detection from our Ellipse R-CNN+ and Ellipse R-CNN on the SOF dataset. While the predicted ellipses from both models are visually comparable, Ellipse R-CNN+ using the KL loss achieves better accuracy as shown in Table~\ref{tab:ue}.}
	\label{fig:experimentUE}
\end{figure}

\begin{table}[!t]
	\caption{Overall performance on the SOF dataset.
		Higher AP is better while lower is better for MR. $\textbf{U}$: ellipse regression with predicted uncertainties.
%		$\text{AP}_{\star}$ and $\text{MR}_{\star}$: the ellipse IoU level starts from 0.7 to 0.9 with an interval 0.05. $\text{AP}_{\star}^{\Theta}$: the angle error decreases from $45^{\circ}$ to $5^{\circ}$ with an interval $10^{\circ}$. The default ellipse IoU for $\text{AP}^{\Theta}$ is 0.7.
		} \label{tab:ue}
	\begin{center}
		\begin{tabular}{l|M{.22cm}|M{.7cm} M{.7cm}|M{.7cm} M{.7cm}|M{.7cm}}
			\toprule
			Methods & \textbf{U} & $\text{AP}_{\star}$ & $\text{AP}_{80}$ & $\text{MR}_{\star}$ & $\text{MR}_{80}$ & $\text{AP}_{\star}^{\Theta}$ \\
			\bottomrule\toprule
			Mask R-CNN+ & -- & 25.7 & 20.3 & 78.3 & 84.1 & 25.6 \\
			Ellipse R-CNN & -- & 41.2 & 39.0 & 63.4 & 68.5 & 56.4 \\
			Ellipse R-CNN+ & \checkmark & \textbf{44.8} & \textbf{42.5} & \textbf{60.0} & \textbf{65.9} & \textbf{60.2} \\
			\bottomrule
		\end{tabular}
	\end{center}
\end{table}

\begin{figure*}[!t]
	\centering
	\includegraphics[width=0.99\textwidth]{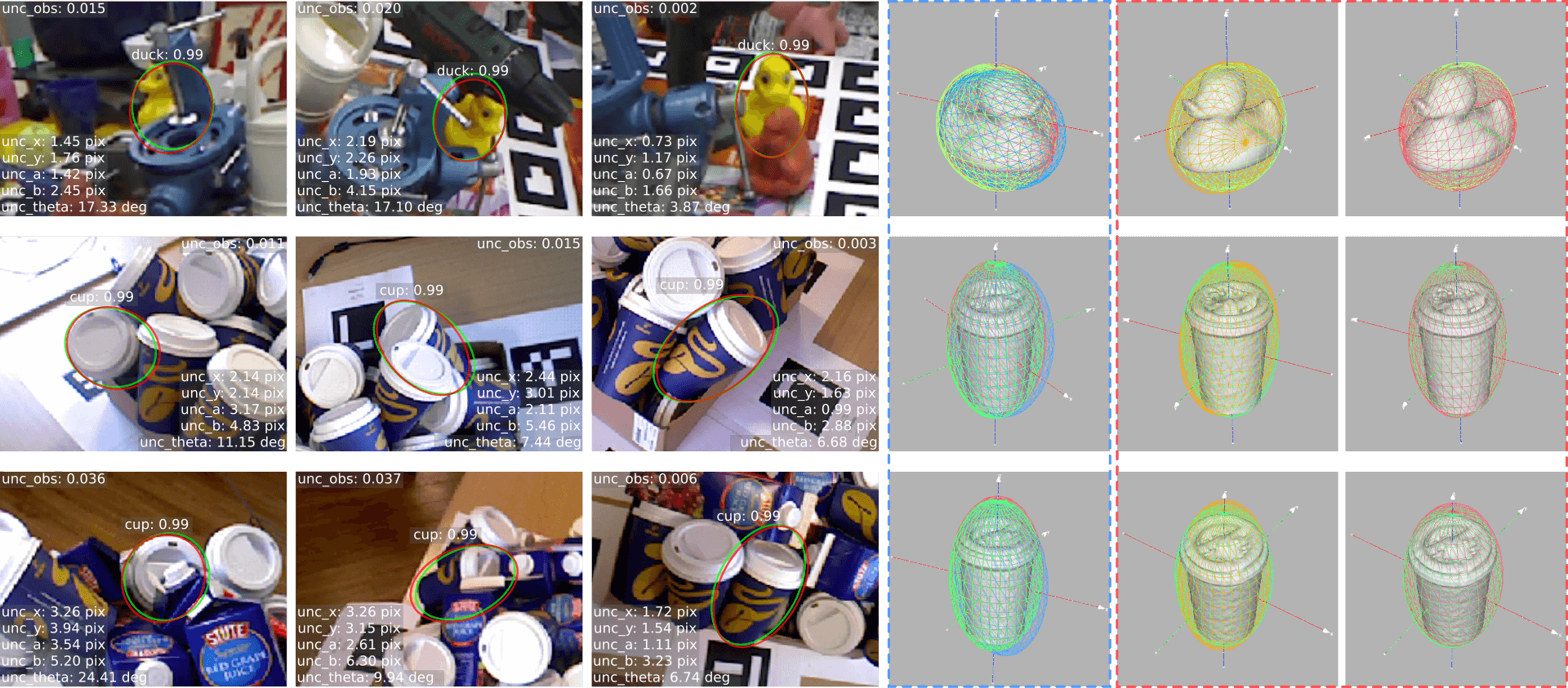}
	\caption{Qualitative results of predicted uncertainties and estimated 3D objects in duck sequence (row 1), cup sequence \texttt{01} (row 2), and sequence \texttt{03} (row 3). From column 3 to column 1, the occlusion level increases. The uncertainties of ellipse parameters are illustrated as \texttt{unc\_x}, \texttt{unc\_y}, \texttt{unc\_a}, and \texttt{unc\_b} in pixels, and \texttt{unc\_theta} in degrees that are transformed from the offsets uncertainties with the predicted visible region (see Eq.~\eqref{eq:normEllipse}). The observation uncertainty \texttt{unc\_obs} indicates the overall geometric quality of each detection. The last three columns show the estimated 3D ellipsoids using different settings: Bbox+ (blue), Ellipse+ (yellow), and EVar+KL (red). In column 4 (blue dashed box), all camera views are used for 3D estimation, while only one-fourth of the views are exploited in the last two columns (red dashed box) to clearly show the difference between Ellipse+ and EVar+KL (proposed). The GT ellipsoids are shown in green. From the comparison, EVar+KL is the most accurate and stable approach to estimating the 3D ellipsoids of objects.}
	\label{fig:expUncertainties}
\end{figure*}

\subsection{Implementation Details}
We implement the Ellipse R-CNN+ model using TensorFlow, and train the networks using a step strategy with mini-batch stochastic gradient descent (SGD) on a GeForce GTX 1080 GPU.
For object detection, we compare our Ellipse R-CNN+ with the original Ellipse R-CNN and Mask R-CNN+ (i.e., Mask R-CNN followed by ellipse fitting~\cite{moshtagh2005minimum, dong2020ellipse}).
All the models are initialized by the pre-trained weights for MS COCO~\cite{lin2014microsoft}.
On SOF, Duck, and Cup datasets, we train the networks with an initial learning rate of $10^{-3}$ for 20,000 iterations and train for another 10,000 iterations with a decreased learning rate of $10^{-4}$.
During training, we perform on-the-fly data augmentation~\cite{sundermeyer2018implicit} with flipping, scaling, shifting, and rotation at random.
We resize the fruit images to $128\times128$, while the duck and cup images are resized to $640\times640$ to keep texture details of higher resolution for training and testing.
For 3D object estimation, we discard almost fully occluded objects (that are not detected by all detectors) in the test sequences of the Cup dataset: 14 cups and 10 cups are available in sequence \texttt{01} and sequence \texttt{03}, respectively.
For the Duck dataset, the duck toy is detectable in all 102 test images for comparison.

\begin{table}[t!]
	\caption{Average $O_\text{3D}$ ($\%$) on Duck and Cup datasets.} \label{tab:O3d}
	\begin{center}
		\begin{tabular}{l l|M{.48cm} M{.48cm} M{.48cm}|M{.48cm} M{.48cm} M{.48cm}}
			\toprule
			\multicolumn{2}{l|}{Datasets} & \multicolumn{3}{c}{Duck}& \multicolumn{3}{c}{Cup} \\
			\multicolumn{2}{l|}{Ratio of camera views} & $1$ & $1/2$ & $1/4$ & $1$ & $1/2$ & $1/4$ \\
			\bottomrule\toprule
			\multirow{4}{*}{\makecell{3D \\ estimation \\ methods}}
			& \multicolumn{1}{|l|}{Bbox+} & 80.5 & 79.0 & 74.6 & 71.0 & 69.8 & 60.1 \\
			& \multicolumn{1}{|l|}{Ellipse+} & 90.5 & 88.5 & 85.3 & 80.6 & 79.8 & 80.2 \\
			\cmidrule{2-8}
			& \multicolumn{1}{|l|}{E+Var} & 92.7 & 92.0 & 90.1 & 83.6 & 83.4 & 84.8 \\
			& \multicolumn{1}{|l|}{EVar+KL} & \textbf{93.6} & \textbf{92.5} & \textbf{91.0} & \textbf{88.0} & \textbf{88.7} & \textbf{89.8} \\
			\bottomrule
		\end{tabular}
	\end{center}
\end{table}

\subsection{Evaluation Metrics}
To evaluate the accuracy of object detection and ellipse regression, we exploit three evaluation metrics: average precision (AP~\cite{lin2014microsoft} over ellipse IoU thresholds), log-average miss rate (MR)~\cite{dollar2011pedestrian}, and $\text{AP}^{\Theta}$ (AP over ellipse angle errors~\cite{dong2020ellipse}).
$\text{AP}^{\Theta}$ focuses more on the accuracy of the predicted ellipse angle: a prediction (evaluated by $\text{AP}_{45}^{\Theta}$) is considered as a false positive if its ellipse IoU is less than 0.7 (the default IoU) or its angle error is greater than $45^{\circ}$.
We apply strict criteria: the IoU level starts from 0.7 up to 0.9 with an interval 0.05 (e.g., average $\text{AP}_{70:90}$ written as $\text{AP}_{\star}$), and the threshold of angle error decreases from $45^{\circ}$ to $5^{\circ}$ with an interval $10^{\circ}$ (i.e., average $\text{AP}_{45:5}^{\Theta}$ written as $\text{AP}_{\star}^{\Theta}$).
For the accuracy of 3D object estimation, three evaluation metrics are $O_\text{3D}$~\cite{rubino20183d}, axis-angle error~\cite{dong2018novel}, and position error.
$O_\text{3D}$ indicates the overall estimation accuracy by measuring the volume intersection over union between GT and estimated ellipsoids in 3D, while the axis-angle error focuses on the angle difference between their major axes in 3D to ignore the inherent model symmetry.
For the test data of the Cup dataset, we calculate the average $O_\text{3D}$ across all estimated cups in both sequences \texttt{01} and \texttt{03}.

\begin{figure*}[!t]
	\centering
	\includegraphics[width=0.99\textwidth]{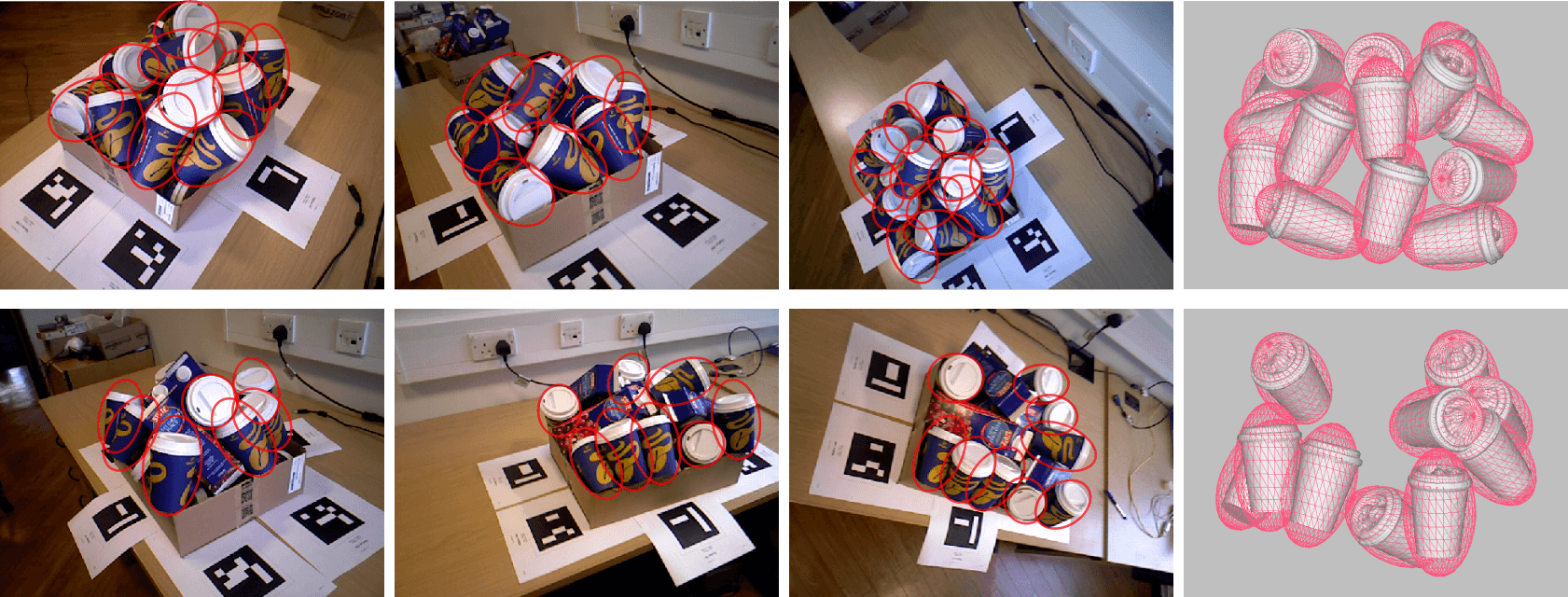}
	\caption{Qualitative results of detected ellipses and estimated 3D ellipsoids of cups using our proposed method (i.e., EVar+KL) for cup sequence \texttt{01} (upper row) and sequence \texttt{03} (lower row). The first three columns show the ellipse detections output by our Ellipse R-CNN+. The last column illustrates the accurate 3D estimation of ellipsoids that well enclose the GT cup models.}
	\label{fig:expDetections}
\end{figure*}

\begin{table*}[thbp!]
	\caption{Average 3D estimation errors (axis-angle error and position error) on Duck and Cup datasets.} \label{tab:anglePos}
	\begin{center}
		\begin{tabular}{l l|M{.48cm} M{.48cm} M{.48cm}|M{.48cm} M{.48cm} M{.48cm}|M{.48cm} M{.48cm} M{.48cm}|M{.48cm} M{.48cm} M{.48cm}}
			\toprule
			\multicolumn{2}{l|}{Datasets} & \multicolumn{6}{c}{Duck} & \multicolumn{6}{c}{Cup} \\
			\multicolumn{2}{l|}{Error metrics} & \multicolumn{3}{c}{Axis-angle error ($^{\circ}$)} & \multicolumn{3}{c|}{Position error (mm)} & \multicolumn{3}{c}{Axis-angle error ($^{\circ}$)}& \multicolumn{3}{c}{Position error (mm)} \\
			\multicolumn{2}{l|}{Ratio of camera views} & $1$ & $1/2$ & $1/4$ & $1$ & $1/2$ & $1/4$ & $1$ & $1/2$ & $1/4$ & $1$ & $1/2$ & $1/4$ \\
			\bottomrule\toprule
			\multirow{4}{*}{\makecell{3D \\ estimation \\ methods}} & \multicolumn{1}{|l|}{Bbox+} & 11.0 & 14.7 & 17.6 & 6.8 & 6.5 & 7.8 & 13.4 & 12.6 & 16.3 & 8.6 & 9.4 & 13.7 \\
			& \multicolumn{1}{|l|}{Ellipse+} & 7.7 & 8.8 & 10.1 & 2.2 & 3.3 & 4.0 & 7.5 & 8.0 & 7.5 & 6.5 & 5.3 & 6.2 \\
			\cmidrule{2-14}
			& \multicolumn{1}{|l|}{E+Var} & 3.4 & \textbf{4.2} & 4.8 & 2.2 & 2.6 & 3.1 & 5.0 & 4.8 & 3.8 & 5.6 & 5.6 & 5.1 \\
			& \multicolumn{1}{|l|}{EVar+KL} & \textbf{3.2} & 4.3 & \textbf{4.6} & \textbf{2.1} & \textbf{2.5} & \textbf{2.8} & \textbf{3.9} & \textbf{3.5} & \textbf{3.7} & \textbf{3.9} & \textbf{3.8} & \textbf{3.0} \\
			\bottomrule
		\end{tabular}
	\end{center}
\end{table*}

\subsection{Comparison Results}
To validate the effectiveness of our predicted uncertainties for accurate 3D object estimation, we compare our probabilistic approach (EVar+KL) with two state-of-the-art methods that use bounding-box constraints~\cite{nicholson2019quadricslam} (Bbox+) and absolute ellipse parameters~\cite{rubino20183d} (Ellipse+), respectively.
In Bbox+, the bounding-box detections of Mask R-CNN+ serve as the inputs for 3D object estimation.
False positives are removed for Mask R-CNN+.
In Ellipse+, we exploit the ellipse detections output by Ellipse R-CNN as the inputs.
We also perform an ablation study of the observation uncertainty, where we only keep the offsets uncertainties in our probabilistic model (i.e., E+Var).

\subsubsection{Detection Accuracy of Ellipse R-CNN+}
We evaluate the contribution of the KL loss in Ellipse R-CNN+.
Table~\ref{tab:ue} shows the detailed breakdown performance.
Some examples of detected occluded objects are illustrated in Fig.~\ref{fig:experimentUE}.
We observe that Ellipse R-CNN+ trained with the KL loss is able to regress ellipses as accurate as Ellipse R-CNN (visually comparable). $\text{AP}_{\star}$, $\text{MR}_{\star}$, and $\text{AP}_{\star}^{\Theta}$ are even further improved by 3.6, 3.4, and 3.8, respectively.
Such improvements are also demonstrated in~\cite{he2019bounding}.
The predicted ellipse offsets with larger differences from their GT have higher uncertainties, which is the extra information to learn.
The KL loss incorporating such regression uncertainties encourages the network to potentially learn more discriminative features for classification.
This gives us a promising way (i.e., more accurate) to deal with the CNN-based regression and classification problems.

\subsubsection{Prediction of Ellipse Uncertainties}
The predicted ellipse uncertainties from our Ellipse R-CNN+ are interpretable.
Fig.~\ref{fig:expUncertainties} shows some qualitative results of uncertainty prediction. 
To clearly illustrate the uncertainties, we transform the offsets uncertainties $\sigma$ into the absolute values in pixels for ellipse location ($x, y$) and size ($a, b$). The uncertainty for ellipse angle $\Theta$ is transformed in degrees.
Such parameter variations are calculated by using Eq.~\eqref{eq:normEllipse} given the output offsets $\delta$, predicted uncertainties, and visible region $Q$.
We observe that the overall geometric quality of each detection is well captured by its observation uncertainty (i.e., \texttt{unc\_obs}).
Specifically, as the occlusion level increases in the duck sequence (the 1st row in Fig.~\ref{fig:expUncertainties}), the 1st view has greatly larger \texttt{unc\_obs} than the 3rd one (0.015 vs. 0.002).
This is because only a small portion of the duck toy is visible such that the predicted ellipse angle has bigger uncertainty ($17.33^\circ$ vs. $3.87^\circ$).
In the cup sequence \texttt{03} (the 3rd row), the uncertainties of ellipse location and size in the 1st view are larger than that in the 3rd view (e.g., 3.54 vs. 1.11 for \texttt{unc\_a}).
The reason is that the image projection of the cup is close to circle and occluded by nearby objects, which is hard to estimate the ellipse confidently.
It is worth noting that the prediction scores of all the detections are almost the same (0.99).
This implies that the classification value is a not good indicator for parameter uncertainties, while our uncertainty model gives an effective way to measure the quality of estimated ellipse parameters, which can be exploited for accurate 3D object estimation.

\subsubsection{Probabilistic 3D Object Estimation}
\begin{figure}[!t]
	\centering
	\includegraphics[width=0.99\columnwidth]{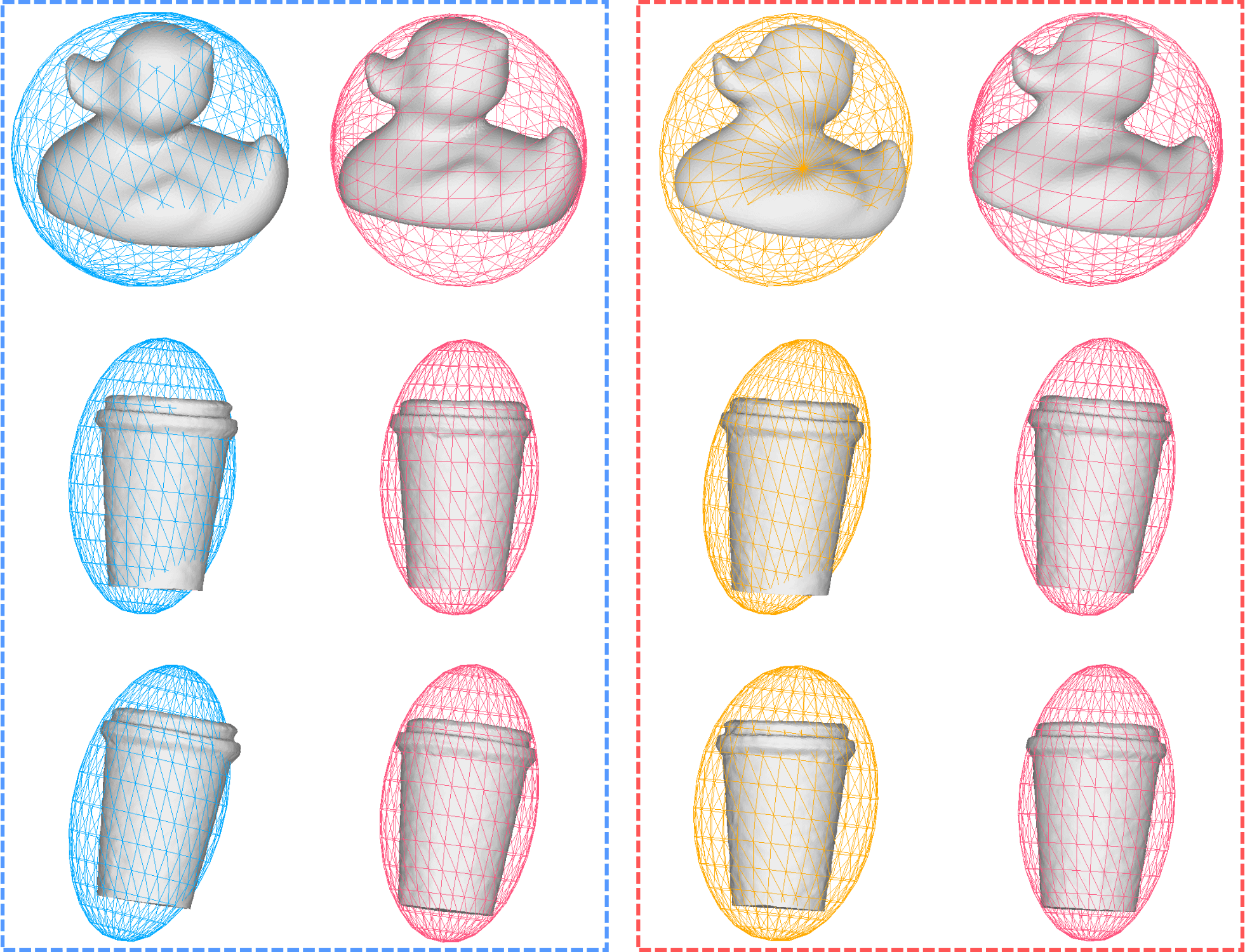}
	\caption{Close-up views of the estimated enclosing ellipsoids using different settings that correspond to the last three columns in Fig.~\ref{fig:expUncertainties}: Bbox+ (blue), Ellipse+ (yellow), and EVar+KL (red). In the blue dashed box, all camera views are used for the comparison between Bbox+ and EVar+KL, while only one-fourth of the views are exploited in the red dashed box to compare Ellipse+ and EVar+KL. In the first row, the estimated ellipsoids from Bbox+ and Ellipse+ cut into the duck meshes, which shows larger position errors from these two methods compared to EVar+KL (ours). In the last two rows, EVar+KL outputs more accurate ellipsoids that better enclose the cup meshes.}
	\label{fig:expCloseup}
\end{figure}

In Fig.~\ref{fig:expUncertainties} (the last three columns) and Fig.~\ref{fig:expCloseup}, some comparison examples of 3D estimation performance for different methods are displayed for duck and cup sequences.
Compared to Bbox+, the estimated ellipsoids output by our EVar+KL almost perfectly fit the GT ones with respect to the size, position, eccentricity, and alignment, as can be seen within the coordinate frame.
To clearly show the difference between Ellipse+ and our method, we only exploit one-fourth of the camera views (the red dashed box in Fig.~\ref{fig:expUncertainties}), and EVar+KL achieves more accurate and stable estimations.
In Table~\ref{tab:O3d}, the overall estimation accuracy of each method on each dataset is reported in terms of $O_\text{3D}$.
In the Duck dataset, due to its large number of views and the good quality of the detections, the average accuracy of Ellipse+ is quite good ($O_\text{3D}$ is 90.5) even without uncertainties integrated.
E+Var and EVar+KL only make a little improvement (92.7 and 93.6).
In the Cup dataset, the limited number of views and the interference from the nearby objects of the same type largely reduce the accuracy of Ellipse+ and Bbox+.
However, integrating the predicted uncertainties greatly improves $O_\text{3D}$, and EVar+KL with the observation uncertainty makes a larger improvement than E+Var (88.0 vs. 83.6). 
Table~\ref{tab:anglePos} also demonstrates such further improvements by Evar+KL in terms of axis-angle and position errors, especially the angle error (i.e., $3.2^\circ$ and $3.9^\circ$ for two datasets).
This is because Ellipse R-CNN+ not only infers the uncertainty of each ellipse offset but also rates the overall quality (i.e., the weight) of each detection from the visible part, which is thus more effective to accurately estimate the 3D size and pose of occluded objects from multiple views.
Fig.~\ref{fig:expDetections} demonstrates the qualitative results of the estimated 3D cup clusters based on multi-view ellipse detections.

To assess the effect of a varying number of camera views, we estimate the 3D ellipsoids of objects using one-fourth, one half, and all of the camera views, respectively.
For each setting, we select the same number of views from each dataset to serve as the same inputs for all methods. 
Tables~\ref{tab:O3d} and~\ref{tab:anglePos} summarize the accuracy evaluation for these three different settings that vary the number of views (see Fig.~\ref{fig:expUncertainties} and Fig.~\ref{fig:expCloseup} for qualitative results).
We observe that the accuracy of Bbox+ degrades significantly in the smaller number of views, while Ellipse+, E+Var, and EVar+KL are affected little and thus insensitive to varying number of views.
This is due to the fact that the geometric constraints of bounding-box detections from multiple views are insufficient and become weaker as the number of views decreases.
Our proposed EVar+KL even improves the accuracy with fewer views, especially for the axis-angle error and $O_\text{3D}$  in the Cup dataset.
The reason is that some views with higher uncertainties may not be selected so that the other views with accurate detections weight more in the 3D estimation.
Even if the estimation results are remarkable for Ellipse+ and E+Var, the average accuracy of EVar+KL is the highest in terms of all the evaluation metrics.

\section{Conclusion} 
In conclusion, traditional object detectors (e.g., Ellipse R-CNN) are not capable of predicting uncertainty information to characterize different occlusion levels for accurate 3D object estimation.
The classification scores do not well indicate the regression uncertainties of the predicted parameters, as shown in our experiments.
In this letter, we propose a novel method of ellipse regression that enables the network to detect the occluded objects and predict their ellipse uncertainties.
Using the KL divergence as the training loss, the deep model learns the ellipse offsets uncertainties and the observation uncertainty that weight each detection accordingly to further boost the 3D estimation accuracy.
Compelling results are demonstrated for Ellipse R-CNN+ on both synthetic and real datasets.
Compared with the non-probabilistic methods, our probabilistic 3D model is able to accurately recover the enclosing ellipsoids (i.e., the size and pose in 3D) of objects even they are heavily occluded from multiple views.
It has also been shown that the performance of our proposed approach is not sensitive to the number of views.
Furthermore, the predicted uncertainties are geometrically interpretable, which may benefit semantic visual SLAM applications.

% or


%\appendices

%\section{}

\section*{Acknowledgment}
We thank our colleagues Cheng Peng, Pravakar Roy, Nicolai H\"{a}ni and Zhihang Deng from the University of Minnesota, for providing valuable feedback and technical support throughout this research.

%\ifCLASSOPTIONcaptionsoff
%  \newpage
%\fi



%\bibliographystyle{IEEEtran}
%\balance
\bibliographystyle{IEEEtran}
\bibliography{ralReferences}

%\begin{thebibliography}{1}
%\end{thebibliography}


\begin{IEEEbiography}{Wenbo Dong:}
Doctor.
\end{IEEEbiography}

%\begin{IEEEbiographynophoto}{Pravakar Roy:}
%Doctor

%\newpage

\begin{IEEEbiographynophoto}{Volkan Isler:}
Professor.
\end{IEEEbiographynophoto}


%\vfill


\end{document}